%% file: main.tex
\pdfoutput=1

\documentclass[11pt]{article}

\usepackage{acl}

\usepackage{times}
\usepackage{latexsym}

\usepackage[T1]{fontenc}

\usepackage[utf8]{inputenc}

\usepackage{microtype}

\usepackage{inconsolata}

\usepackage{amsmath,amsfonts,amstext}
\usepackage{tikz}
\usetikzlibrary{positioning,calc,patterns}
\usepackage{array}
\usepackage{booktabs}
\usepackage{diagbox}
\usepackage{standalone}
\usepackage{subcaption}
\usepackage{ezutils}

\newcommand{\avestd}[2]{#1\textsubscript{\textpm #2}}
\newcommand{\gstyle}[1]{\textit{\textcolor{gray}{#1}}}
\newcommand{\gavestd}[2]{\gstyle{#1}\textsubscript{\gstyle{\textpm #2}}}

\newcommand{\ann}[3]{\textcolor{#1}{\textsf{[}}#3\textcolor{#1}{\textsf{]}\textsubscript{\texttt{#2}}}}
\newcommand{\PER}[1]{\ann{casgreen}{PER}{#1}}
\newcommand{\ORG}[1]{\ann{casgreen}{ORG}{#1}}
\newcommand{\LOC}[1]{\ann{casgreen}{LOC}{#1}}
\newcommand{\MISC}[1]{\ann{casgreen}{MISC}{#1}}

\newcommand{\mORG}[1]{\ann{casred}{ORG}{#1}}
\newcommand{\mLOC}[1]{\ann{casred}{LOC}{#1}}

\newcommand{\ePER}[1]{\ann{casyellow}{PER}{#1}}
\newcommand{\eORG}[1]{\ann{casyellow}{ORG}{#1}}
\newcommand{\eLOC}[1]{\ann{casyellow}{LOC}{#1}}
\newcommand{\eMISC}[1]{\ann{casyellow}{MISC}{#1}}

\title{Recognizing Nested Entities from Flat Supervision: A New NER Subtask, Feasibility and Challenges}

\author{Enwei Zhu \and Yiyang Liu \and Ming Jin \and Jinpeng Li\textsuperscript{\footnotemark[1]}  \\
        HwaMei Hospital, University of Chinese Academy of Sciences \\
        Ningbo Institute of Life and Health Industry, University of Chinese Academy of Sciences \\ 
    	\texttt{\{zhuenwei,liuyiyang,jinming,lijinpeng\}@ucas.ac.cn}}

\begin{document}
\maketitle
{\renewcommand{\thefootnote}{\fnsymbol{footnote}}
\footnotetext[1]{Corresponding author.}
}

\begin{abstract}
Many recent named entity recognition (NER) studies criticize flat NER for its non-overlapping assumption, and switch to investigating nested NER. However, existing nested NER models heavily rely on training data annotated with nested entities, while labeling such data is costly. This study proposes a new subtask, \emph{nested-from-flat NER}, which corresponds to a realistic application scenario: given data annotated with flat entities only, one may still desire the trained model capable of recognizing nested entities. 

To address this task, we train span-based models and deliberately ignore the spans nested inside labeled entities, since these spans are possibly unlabeled entities. With nested entities removed from the training data, our model achieves 54.8\%, 54.2\% and 41.1\% $F_1$ scores on the subset of spans within entities on ACE 2004, ACE 2005 and GENIA, respectively. This suggests the effectiveness of our approach and the feasibility of the task. In addition, the model's performance on flat entities is entirely unaffected. We further manually annotate the nested entities in the test set of CoNLL 2003, creating a nested-from-flat NER benchmark.\footnote{Our code and annotations will be publicly released.} Analysis results show that the main challenges stem from the data and annotation inconsistencies between the flat and nested entities. 
\end{abstract}

\section{Introduction}
Named entity recognition (NER) is a fundamental natural language processing (NLP) task that requires detecting text spans of interest, and classifying them into pre-defined entity categories, e.g., Person, Organization, Location. Researchers had been long-term investigating \emph{flat NER} where entity spans are assumed non-overlapping~\citep{collobert2011natural, huang2015bidirectional, lample-etal-2016-neural}, while many recent studies criticize such flat setting and switch to \emph{nested NER} that allows an entity to contain other entities inside it~\citep{katiyar-cardie-2018-nested, sohrab-miwa-2018-deep, yu-etal-2020-named, yan-etal-2021-unified-generative}. For example, a location entity ``New York'' can be nested in an organization entity ``New York University''. In nested NER, the two entities are equally considered from annotation through evaluation, while flat NER focuses on the outer entity but ignores the nested one entirely~\citep{tjong-kim-sang-de-meulder-2003-introduction, finkel-manning-2009-nested}.\footnote{See CoNLL 2003 Annotation Guidelines (\url{https://www-nlpir.nist.gov/related_projects/muc/proceedings/ne_task.html}), Subsections~4.3 and A.1.3.} Nested NER appears a more general and realistic setting since nested entities are ubiquitous in natural language. 

Labeling nested entities is particularly labor-intensive, complicated, and error-prone; for example, \citet{ringland-etal-2019-nne} reported that entities can be nested up to six layers. However, all existing nested NER systems heavily rely on \emph{nested supervision}, namely training on annotated nested NER datasets, such as ACE 2004, ACE 2005 and GENIA~\citep{kim2003genia}. Directly imposing \emph{flat supervision} would misguide the models to ignore nested structures. This creates an obstacle for them to utilize well-annotated flat NER resources, such as CoNLL 2003~\citep{tjong-kim-sang-de-meulder-2003-introduction} and OntoNotes 5. 

This study proposes a new subtask, \emph{nested-from-flat NER}, which asks to train a nested NER model with purely flat supervision. This corresponds to a realistic application scenario: given training data annotated with flat entities only, one may still desire the trained model capable of extracting nested entities from unseen text. 

To address this challenging task, we exploit the span-based NER framework which explicitly distinguishes positive samples (i.e., entity spans) from negative samples (i.e., non-entity spans). When training a span-based neural NER model, a standard protocol regards all unannotated spans as negative samples~\citep{sohrab-miwa-2018-deep, yu-etal-2020-named, zhu2022deep}; however, we deliberately ignore the spans nested in any labeled entities, because these spans are possibly unlabeled nested entities. When the trained model generalizes to all spans, it naturally predicts all possible entities, which may contain nested ones. This is theoretically feasible because the recognizable patterns for flat entities should be to some extent transferable to nested entities. 

With nested entities removed from the training splits of ACE 2004, ACE 2005 and GENIA, the nested-from-flat model can achieve 54.8\%, 54.2\% and 41.1\% $F_1$ scores on the subset of spans within entities, respectively. Besides, the overall $F_1$ scores reach 79.2\%, 79.3\% and 77.3\%. Moreover, the model's ability to recognize nested entities does not hurt its performance on flat entities. We further annotate the nested entities in the test split of CoNLL 2003, and analyze the recognition results of our models. We find nested-from-flat NER a challenging task mainly because the annotation standards and data distributions are inconsistent between the flat and nested entities. 

This study contributes in threefold:
\begin{itemize}
    \item We propose nested-from-flat NER, a new subtask with realistic application scenarios. Compatibly, we design a metric -- $F_1$ score on the spans within entities, which dedicatedly evaluates how well the model extracts nested entities. 
    \item We provide a solution to nested-from-flat NER, which simply ignores the spans nested in entities during training. Experimental results confirm its effectiveness, as well as the feasibility of this task. 
    \item We manually annotate the nested entities in the test split of CoNLL 2003, resulting in a nested-from-flat NER benchmark named CoNLL 2003 NFF. 
\end{itemize}

\section{Related Work}
The NER task was originally proposed in a context where entities could be regarded as small chunks and thus detected by finite state models~\citep{finkel-manning-2009-nested}. Hence, in the early years, NER corpus designers chose to annotate only the outermost entities, but ignore/remove the nested ones~\citep{tjong-kim-sang-de-meulder-2003-introduction, collier2004introduction}; and algorithm researchers were focused on using sequence models, such as the conditional random field (CRF)~\citep{lafferty2001conditional},
to recognize flat entities. Facilitated by the deep learning technologies~\citep{krizhevsky2012imagenet, lecun2015deep}, neural sequence tagging models with an optional linear-chain CRF became the \emph{de facto} standard solution to flat NER~\citep{collobert2011natural, huang2015bidirectional, lample-etal-2016-neural, zhang-yang-2018-chinese, devlin-etal-2019-bert}. 

However, nested entities are ubiquitous in natural language. Many recent studies criticize the flat assumption, and switch to a setting that allows nested entities~\citep{finkel-manning-2009-nested}. This also remarkably facilitates the progress in NER system designs beyond the traditional sequence tagging framework. 
Hypergraph-based models adopt a tagging scheme that allows multiple tags for a single token and multiple transitions between tags at adjacent positions, and thus complies with nested structures~\citep{lu-roth-2015-joint, katiyar-cardie-2018-nested}. Span-based methods enumerate or propose candidate spans, and then classify the spans into entity categories~\citep{sohrab-miwa-2018-deep, eberts2020span, yu-etal-2020-named, shen-etal-2021-locate}. Other approaches include stacked sequence tagging models~\citep{ju-etal-2018-neural}, reformulating NER as a reading comprehension task~\citep{li-etal-2020-unified} or a generation task~\citep{yan-etal-2021-unified-generative}, set prediction~\citep{tan2021sequence, shen-etal-2022-parallel}, and word-word relation prediction~\citep{li2022unified}. 

Almost all the existing nested NER models heavily rely on annotated nested NER resources, while labeling nested entities is labor-intensive, complicated and error-prone~\citep{ringland-etal-2019-nne}. This study proposes nested-from-flat NER, exploring the possibility of training a nested NER model with flatly annotated data. 

\section{Method}
\paragraph{Span-based NER.} Given a $T$-length sentence, a span-based neural NER model enumerates all possible spans, and builds a span representation $\ve{z}_{ij} \in \mathbb{R}^d$ for each span $(i, j)$, typically based on the contextualized embeddings from a pretrained language model (PLM). The span representations are then fed into a classifier: 
\begin{equation}
\hat{\ve{y}}_{ij} = \softmax (\ma{W} \ve{z}_{ij} + \ve{b}), 
\end{equation}
where $\ma{W} \in \mathbb{R}^{c \times d}$ and $\ve{b} \in \mathbb{R}^c$ are learnable parameters, and $\hat{\ve{y}}_{ij} \in \mathbb{R}^c$ is the estimated posterior probabilities over entity types (including an additional ``non-entity'' type). 

Given the one-hot encoded ground truth $\ve{y}_{ij} \in \mathbb{R}^c$, the model can be trained by optimizing the cross-entropy loss for all spans: 
\begin{equation}
\mathcal{L} = -\sum_{0 \leq i \leq j < T} \ve{y}_{ij}^\trans \log(\hat{\ve{y}}_{ij}). 
\end{equation}

In the inference time, the spans predicted to be ``non-entity'' are discarded; while the remaining ones, together with their predicted types, are output as recognized entities. 

\paragraph{Nested-from-Flat NER.} We start from an example sentence: ``Mr. John Smith graduated from New York University last year''. In a typical nested NER annotation scheme, ``John Smith'', ``New York University'' and ``New York'' should be labeled as Person, Organization and Location entities, respectively; while flat NER ignores any nested entities, namely ``New York'' in this example. Nested-from-flat NER asks: if only flat entities are available in the training data, how to develop a model that recognizes nested entities in unseen sentences? 

Formally, given a sentence annotated with flat entities, denote all the spans as a set $\mathcal{A}$, and the entity spans as a set $\mathcal{E}$. A standard span-based NER modeling protocol regards $\mathcal{E}$ as positive samples, and $\mathcal{A} \setminus \mathcal{E}$ as negative samples. However, unlabeled nested entities may exist in $\mathcal{A} \setminus \mathcal{E}$ and thus be incorrectly treated as negative samples. 

\begin{figure}[t]
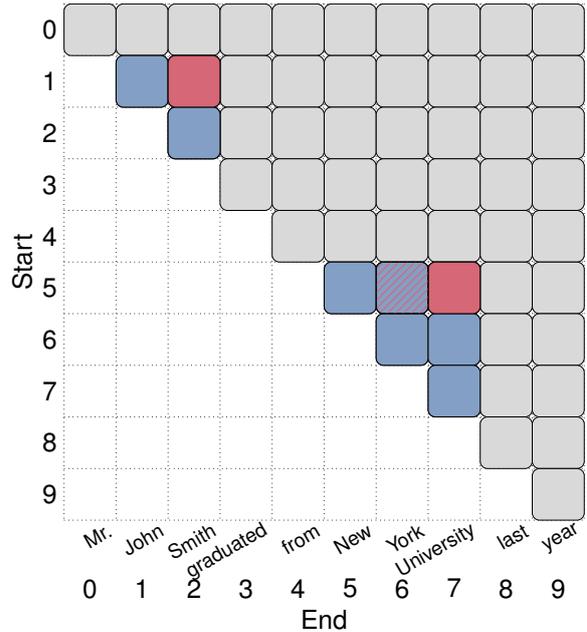

    \centering
    \includestandalone[width=\columnwidth]{tikz/inside-and-outside-spans}
    \caption{Within-entity and out-of-entity spans for sentence ``Mr. John Smith graduated from New York University last year''. 
    Within-entity spans are colored in \textcolor{casblue}{blue}. 
    Out-of-entity spans include the entity spans colored in \textcolor{casred}{red}, and those colored in \textcolor{gray}{gray}. 
    The span colored in \textcolor{casblue}{blue} but hatched in \textcolor{casred}{red} is an unlabeled nested entity.}
    \label{fig:inside-and-outside-spans}
\end{figure}

To address this issue, we define the set of \emph{within-entity spans}: 
\begin{equation}
\begin{aligned}
\mathcal{I} = \{(s, e) \mid \exists\ (s', e') \in \mathcal{E}, \text{s.t.} \, & s' \leq s \leq e < e' \\ 
\text{or} \, & s' < s \leq e \leq e' \},  
\end{aligned}
\end{equation}
and the set of \emph{out-of-entity spans}: 
\begin{equation}
\mathcal{O} = \mathcal{A} \setminus \mathcal{I}. 
\end{equation}
Note that the two sets are mutually exclusive; and the entity spans belong to the out-of-entity spans, i.e., $\mathcal{E} \subseteq \mathcal{O}$. Figure~\ref{fig:inside-and-outside-spans} visualizes the two sets of spans of the aforementioned 10-token sentence, which cover the upper triangular area of the resulting 10x10 matrix. 

Clearly, unlabeled nested entities can only appear in the within-entity spans $\mathcal{I}$, rather than the out-of-entity spans $\mathcal{O}$. Therefore, the supervisory signals (positive vs. negative samples) are reliable in $\mathcal{O}$, but of high risk in $\mathcal{I}$. 

This leads to the key ingredient of our solution. In the training time, we train the model with samples from $\mathcal{O}$ while ignore $\mathcal{I}$: 
\begin{equation}
\mathcal{L} = -\sum_{(i, j) \in \mathcal{O}} \ve{y}_{ij}^\trans \log(\hat{\ve{y}}_{ij}). 
\end{equation}
Empirically, $\mathcal{O}$ contains substantially more span-level samples than $\mathcal{I}$. For example, the out-of-entity spans are over 100 times more than the within-entity spans in the CoNLL 2003 training split. Hence, the out-of-entity spans are sufficient for training the model. 

In the inference time, we let the model generalize to all spans of test sentences, predicting all possible entities, regardless nested or not. If the model is well-trained, it is able to recognize entities nested within others.

\paragraph{Negative Sampling on Within-Entity Spans.} As aforementioned, in a standard protocol, the within-entity spans $\mathcal{I}$ are all regarded as negative samples because they are unlabeled. While in our method for nested-from-flat NER, the spans in $\mathcal{I}$ are entirely ignored in training. 

Inspired by \citet{li2021empirical}, we find it sometimes beneficial to additionally sample a few spans from $\mathcal{I}$ and use them as negative samples. To formulate this trick, we introduce a hyperparameter $\gamma$, which represents the negative sampling rate for $\mathcal{I}$. Thus, we have three schemes for the within-entity spans: 
\begin{itemize}
    \item \emph{Full Negative} ($\gamma = 1$): Using all spans in $\mathcal{I}$ as negative samples; this corresponds to the standard span-based NER training protocol. 
    \item \emph{Sampling} ($0 < \gamma < 1$): Randomly sampling spans with probability $\gamma$ from $\mathcal{I}$ as negative samples. Empirically, a relatively small sampling rate (e.g., $\gamma = 0.01$, our default) works much better than large rates. 
    \item \emph{Full Ignoring} ($\gamma = 0$): Ignoring $\mathcal{I}$ in training. 
\end{itemize}

\section{Experimental Settings}
\paragraph{Datasets.} Although our model can be trained on flatly annotated data, it relies on test data with nested entities to evaluate the trained model. Hence, we first perform experiments on some nested NER benchmarks, i.e., ACE 2004\footnote{\url{https://catalog.ldc.upenn.edu/LDC2005T09}.}, ACE 2005\footnote{\url{https://catalog.ldc.upenn.edu/LDC2006T06}.} and GENIA~\citep{kim2003genia}. Before training, we deliberately remove all the nested entities but keep the outermost ones in the training and development splits~\citep{finkel-manning-2009-nested}, which thus satisfies the nested-from-flat setting. 

In addition, we manually annotate the nested entities in the test split of CoNLL 2003~\citep{tjong-kim-sang-de-meulder-2003-introduction}. Three human experts were hired for this project, and they were asked to strictly follow the original CoNLL 2003 Annotation Guidelines. The final annotation results are based on an additional round of manual validation that resolves the disagreements between the annotators. The resulting dataset is named \emph{CoNLL 2003 NFF}, which dedicatedly serves as a nested-from-flat NER benchmark. More details on the data annotation, processing and descriptive statistics can be found in Appendix~\ref{app:datasets}. 

\paragraph{Evaluation.} Same as in the standard NER, an entity is evaluated to be correct if both its predicted boundaries and category exactly match the ground truth. The evaluation metric is the micro $F_1$ score on the test split. Unless otherwise noted, we run each experiment for 10 times and report the average $F_1$ score with corresponding standard deviation. 

In addition to the overall $F_1$ score that considers all spans, we separately evaluate the trained model on the within-entity spans $\mathcal{I}$ and the out-of-entity spans $\mathcal{O}$, yielding within-entity and out-of-entity $F_1$ scores, respectively. In this study, the within-entity $F_1$ score is the core metric, which reflects how well the model recognizes nested entities. 

\paragraph{Hyperparameters.} In all experiments, we use RoBERTa~\citep{liu2019roberta} of the \texttt{base} size (12 layers, 768 hidden size) as the PLM, followed by a single-layer 400-dimensional LSTM~\citep{hochreiter1997long}. We choose three representative span-based NER decoders, i.e., SpERT~\citep{eberts2020span}, biaffine~\cite{yu-etal-2020-named} and DSpERT~\citep{zhu2022deep}. In addition, boundary smoothing regularization~\citep{zhu-li-2022-boundary} is applied with $\epsilon = 0.1$. 

The models are trained by the AdamW optimizer~\citep{loshchilov2018decoupled} for 20 epochs with batch size 48. Gradients are clipped at $\ell_2$-norm of 5~\citep{pascanu2013difficulty}. The learning rates are 1.5e-5 and 2.5e-3 for pretrained weights and randomly initialized weights, respectively; a scheduler of linear warmup is applied in the first 20\% epochs followed by linear decay.

\section{Results on Nested NER Datasets}
Table~\ref{tab:nested-res} presents the evaluation results, i.e., within-entity, out-of-entity and overall $F_1$ scores of three span-based models on ACE 2004, ACE 2005 and GENIA. \emph{Full Negative}, \emph{Sampling} and \emph{Full Ignoring} are three training schemes described above, which perform nested-from-flat experiments where the nested entities are removed from the training and development splits. \emph{Sampling} uses a fixed rate $\gamma = 0.01$. An exception is \emph{Gold Superv.}, which retains and uses the ground-truth nested entities for training; this serves as an empirical upper bound for the nested-from-flat results. 

\begin{table*}[ht]
    \centering \small
    \begin{tabular}{lccccccccc}
        \toprule
        & \multicolumn{3}{c}{ACE 2004} & \multicolumn{3}{c}{ACE 2005} & \multicolumn{3}{c}{GENIA} \\
         \cmidrule(lr){2-4} \cmidrule(lr){5-7} \cmidrule(lr){8-10} 
         & Within & Out & Overall & Within & Out & Overall & Within & Out & Overall \\
        \midrule
        SpERT \\
        \quad + Full Negative         & \avestd{~~7.7}{0.4} & \avestd{84.1}{0.9} & \avestd{69.6}{0.8} & \avestd{~~7.6}{1.6} & \avestd{82.1}{0.7} & \avestd{71.0}{0.6} & \avestd{~~7.5}{0.9} & \avestd{79.6}{0.7} & \avestd{74.3}{0.7} \\
        \quad + Sampling              & \avestd{17.3}{1.1} & \avestd{\textbf{86.3}}{0.4} & \avestd{\textbf{71.4}}{0.6} & \avestd{23.8}{2.0} & \avestd{\textbf{84.8}}{0.2} & \avestd{\textbf{72.9}}{0.5} & \avestd{26.7}{1.8} & \avestd{\textbf{80.8}}{0.5} & \avestd{\textbf{74.8}}{0.5} \\
        \quad + Full Ignoring         & \avestd{\textbf{21.2}}{1.1} & \avestd{\textbf{86.4}}{0.5} & \avestd{65.0}{1.1} & \avestd{\textbf{27.2}}{0.7} & \avestd{\textbf{84.7}}{0.4} & \avestd{69.1}{0.8} & \avestd{\textbf{28.2}}{1.8} & \avestd{\textbf{80.5}}{0.3} & \avestd{74.4}{0.3} \\
        \quad + \gstyle{Gold Superv.} & \gavestd{77.2}{1.3} & \gavestd{84.5}{0.8} & \gavestd{82.3}{0.7} & \gavestd{73.8}{0.9} & \gavestd{83.2}{0.2} & \gavestd{80.9}{0.3} & \gavestd{51.9}{0.5} & \gavestd{81.0}{0.5} & \gavestd{77.7}{0.4} \\
        \midrule
        Biaffine \\
        \quad + Full Negative         & \avestd{~~9.1}{0.4} & \avestd{86.9}{0.3} & \avestd{72.3}{0.2} & \avestd{~~9.9}{1.3} & \avestd{84.4}{0.4} & \avestd{73.5}{0.3} & \avestd{15.4}{1.1} & \avestd{82.6}{0.2} & \avestd{\textbf{77.1}}{0.2} \\
        \quad + Sampling              & \avestd{34.0}{2.1} & \avestd{\textbf{88.0}}{0.3} & \avestd{\textbf{74.9}}{0.5} & \avestd{41.2}{1.3} & \avestd{\textbf{86.1}}{0.3} & \avestd{\textbf{77.0}}{0.3} & \avestd{\textbf{39.4}}{1.1} & \avestd{\textbf{83.7}}{0.3} & \avestd{76.6}{0.2} \\
        \quad + Full Ignoring         & \avestd{\textbf{40.9}}{1.3} & \avestd{\textbf{88.1}}{0.2} & \avestd{74.4}{0.4} & \avestd{\textbf{45.4}}{1.1} & \avestd{\textbf{86.3}}{0.4} & \avestd{76.7}{0.4} & \avestd{\textbf{39.1}}{1.0} & \avestd{\textbf{83.8}}{0.3} & \avestd{76.2}{0.3} \\
        \quad + \gstyle{Gold Superv.} & \gavestd{86.2}{0.2} & \gavestd{87.5}{0.2} & \gavestd{87.1}{0.2} & \gavestd{83.7}{0.6} & \gavestd{85.9}{0.2} & \gavestd{85.4}{0.3} & \gavestd{54.2}{0.6} & \gavestd{83.4}{0.2} & \gavestd{79.6}{0.1} \\
        \midrule
        DSpERT \\
        \quad + Full Negative         & \avestd{~~9.4}{0.7} & \avestd{86.9}{0.2} & \avestd{72.3}{0.2} & \avestd{11.1}{1.6} & \avestd{84.5}{0.2} & \avestd{73.7}{0.2} & \avestd{10.3}{0.7} & \avestd{82.9}{0.3} & \avestd{\textbf{77.3}}{0.3} \\
        \quad + Sampling              & \avestd{\textbf{54.8}}{1.3} & \avestd{\textbf{88.6}}{0.1} & \avestd{\textbf{79.2}}{0.4} & \avestd{\textbf{54.2}}{1.2} & \avestd{\textbf{86.7}}{0.2} & \avestd{\textbf{79.3}}{0.2} & \avestd{\textbf{41.1}}{0.9} & \avestd{\textbf{83.7}}{0.6} & \avestd{76.6}{0.5} \\
        \quad + Full Ignoring         & \avestd{39.4}{2.5} & \avestd{85.9}{1.4} & \avestd{65.6}{2.0} & \avestd{39.6}{3.6} & \avestd{84.9}{1.3} & \avestd{68.5}{2.6} & \avestd{\textbf{40.9}}{1.1} & \avestd{\textbf{83.3}}{0.7} & \avestd{76.0}{0.4} \\
        \quad + \gstyle{Gold Superv.} & \gavestd{87.0}{0.3} & \gavestd{88.0}{0.3} & \gavestd{87.7}{0.2} & \gavestd{85.6}{0.6} & \gavestd{86.0}{0.2} & \gavestd{85.9}{0.2} & \gavestd{55.9}{0.8} & \gavestd{83.7}{0.1} & \gavestd{80.3}{0.1} \\
        \bottomrule
    \end{tabular}
    \caption{Results of nested-from-flat experiments on nested NER datasets. Reported are average $F_1$ scores with corresponding standard deviations of 10 independent runs. 
    The normally styled rows are results by a nested-from-flat setting where nested entities are removed from the training and development splits. 
    The \gstyle{gray italicized} rows (i.e., ``\gstyle{Gold Superv.}'') are results with nested entities retained and used in training; these serve as an empirical upper bound for the nested-from-flat experiments.
    The best $F_1$ scores are in bold for each model.}
    \label{tab:nested-res}
\end{table*}

\begin{figure*}[ht]
    \centering
    \begin{subfigure}{0.32\textwidth}
    \centering
    \includegraphics[width=\textwidth]{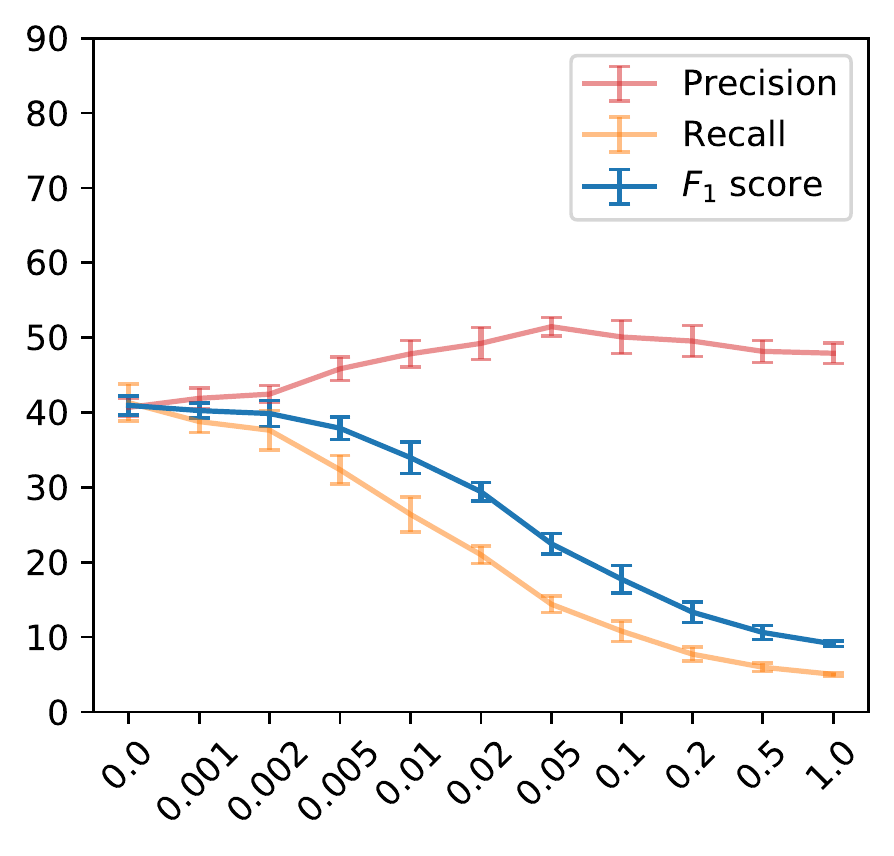}
    \caption{Biaffine, ACE 2004} \label{subfig:biaffine-on-ace04}
    \end{subfigure}
    \begin{subfigure}{0.32\textwidth}
    \centering
    \includegraphics[width=\textwidth]{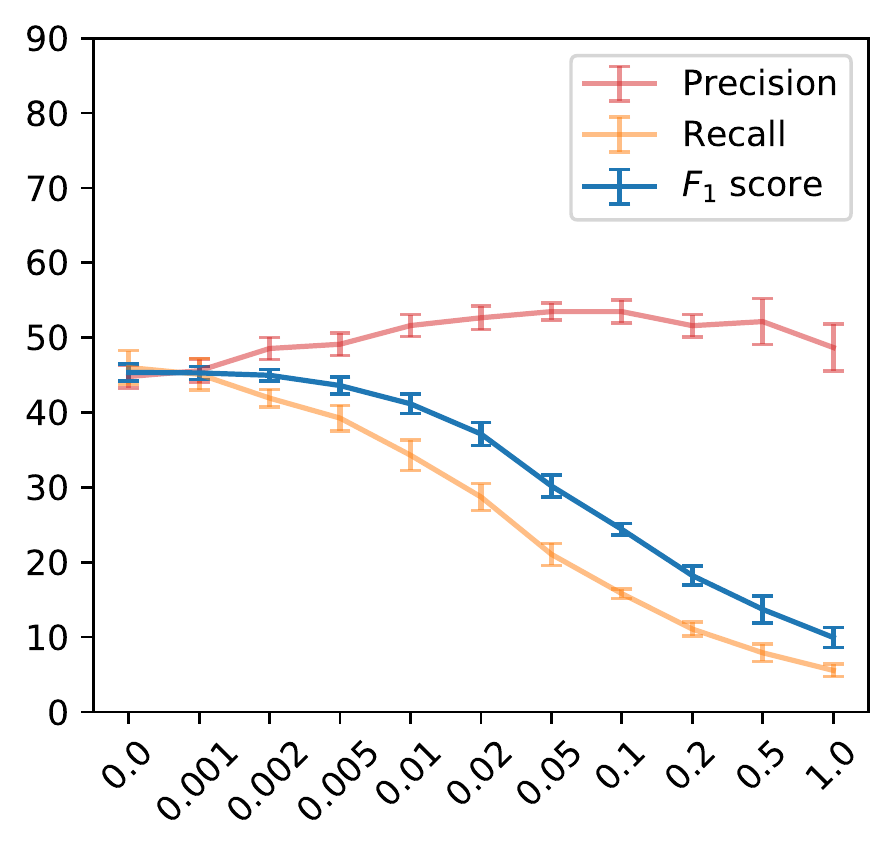}
    \caption{Biaffine, ACE 2005} \label{subfig:biaffine-on-ace05}
    \end{subfigure}
    \begin{subfigure}{0.32\textwidth}
    \centering
    \includegraphics[width=\textwidth]{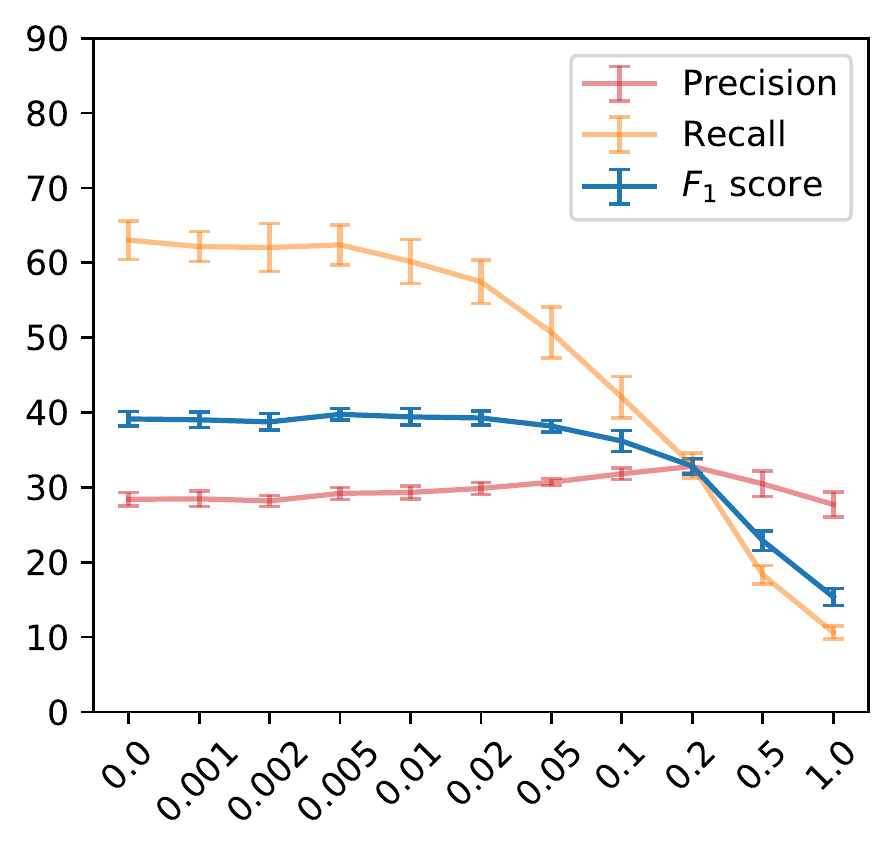}
    \caption{Biaffine, GENIA} \label{subfig:biaffine-on-genia}
    \end{subfigure}
    \begin{subfigure}{0.32\textwidth}
    \centering
    \includegraphics[width=\textwidth]{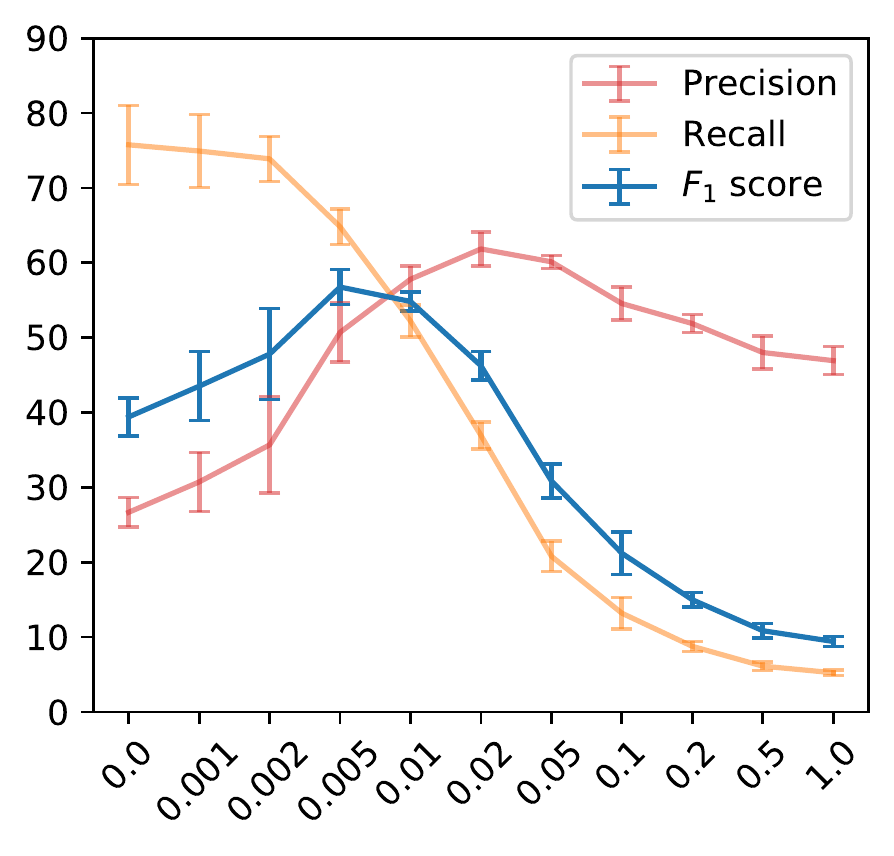}
    \caption{DSpERT, ACE 2004} \label{subfig:dspert-on-ace04}
    \end{subfigure}
    \begin{subfigure}{0.32\textwidth}
    \centering
    \includegraphics[width=\textwidth]{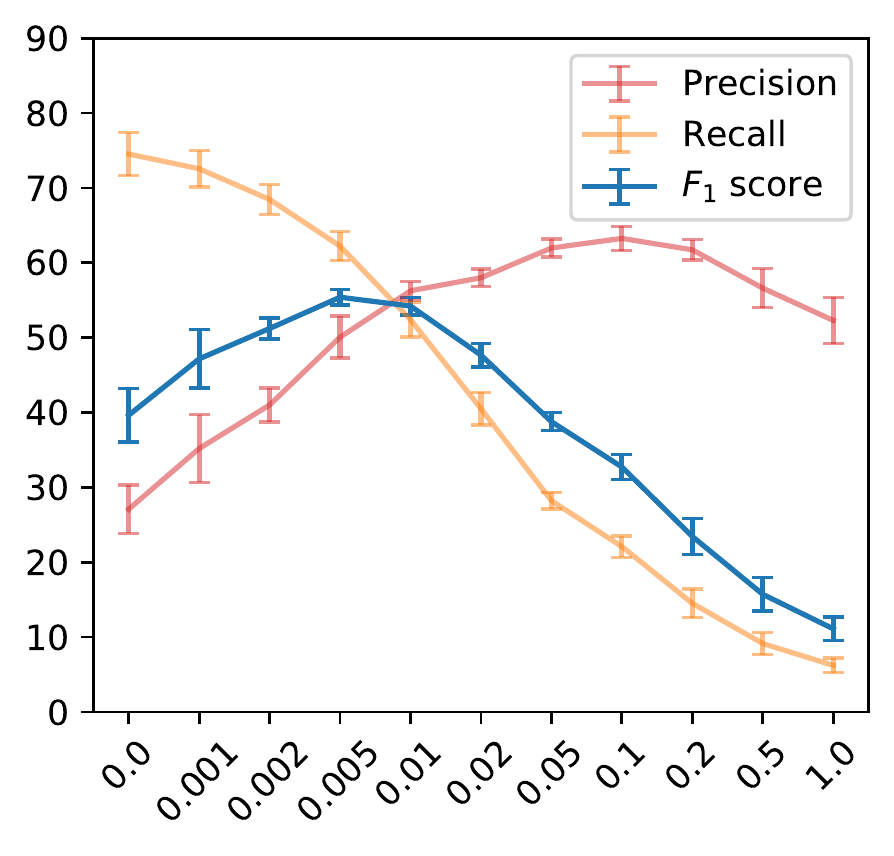}
    \caption{DSpERT, ACE 2005} \label{subfig:dspert-on-ace05}
    \end{subfigure}
    \begin{subfigure}{0.32\textwidth}
    \centering
    \includegraphics[width=\textwidth]{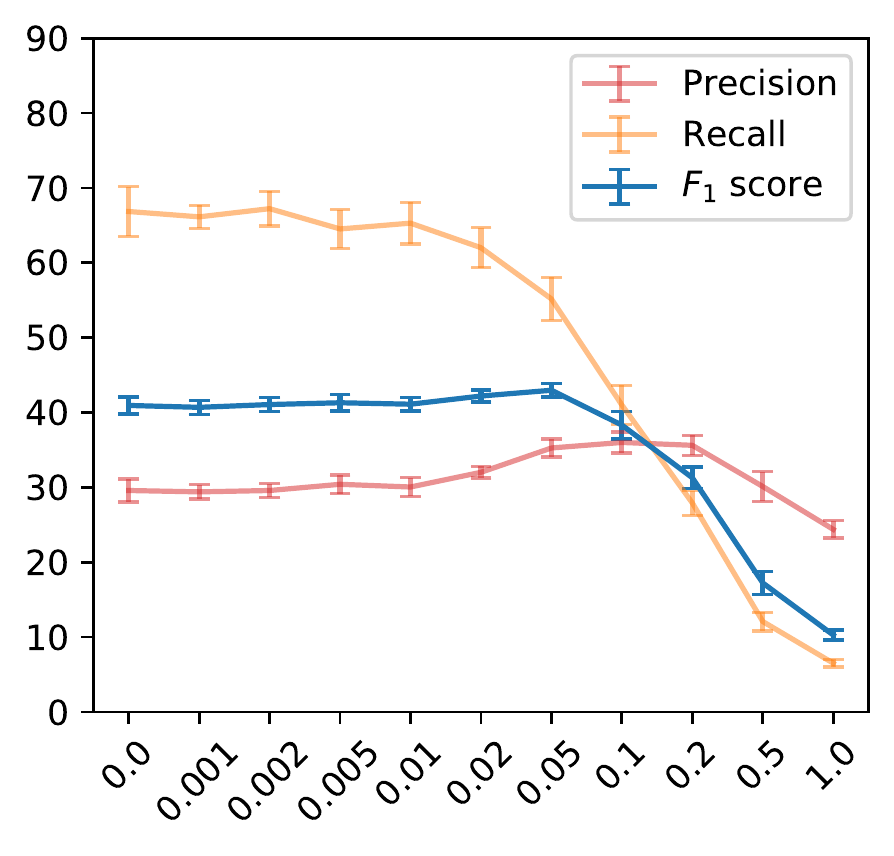}
    \caption{DSpERT, GENIA} \label{subfig:dspert-on-genia}
    \end{subfigure}
    \caption{Precision, recall and $F_1$ scores on within-entity spans by different sampling rates. All the results are average scores of 10 independent runs; the error bars represent the corresponding standard deviations.}
    \label{fig:f1-by-nsr}
\end{figure*}

DSpERT+\emph{Sampling} appears the best configuration for recognizing nested entities, achieving within-entity $F_1$ scores of 54.8\%, 54.2\% and 41.1\% on ACE 2004, ACE 2005 and GENIA, respectively. These scores are largely 2/3 -- 3/4 of the corresponding ``oracle'' results (i.e., 87.0\%, 85.6\% and 55.9\%) by gold supervision. Considering the unavailability of nested supervision, such performance is very encouraging, suggesting the feasibility of the nested-from-flat NER task. In addition, \emph{Full Ignoring} and \emph{Sampling} significantly outperform the standard span-based NER training protocol, i.e., \emph{Full Negative}, across all models and datasets; this suggests the effectiveness of our proposed approach. As previously analyzed, the within-entity spans probably contain unlabeled nested entities, so treating them all as negative samples strongly biases the model's behavior in recognizing nested entities. 

For each model, the out-of-entity $F_1$ scores are in general of similar magnitudes across different schemes. This means that the additional ability for recognizing nested entities is obtained for free, without any performance sacrifice on the flat (outermost) entities. 

The best overall $F_1$ scores by the nested-from-flat models are 79.2\%, 79.3\% and 77.3\% on ACE 2004, ACE 2005 and GENIA, respectively. Such performance is also competitive, with 3.0 -- 8.5 percentage gaps to the upper bounds. Note that our results without any nested supervision are even comparable to the state-of-the-art nested NER performance reported several years ago~\citep[e.g.,][]{katiyar-cardie-2018-nested, wang-lu-2018-neural}. 

\paragraph{Effect of Negative Sampling.} According to the experimental results, DSpERT perform best with \emph{Sampling} (i.e., $\gamma = 0.01$), while SpERT and biaffine are more compatible with \emph{Full Ignoring} (i.e., $\gamma = 0$). To investigate the effect of negative sampling, we plot the within-entity precision rates, recall rates and $F_1$ scores for different negative sampling rates in Figure~\ref{fig:f1-by-nsr}. It shows that the resulting patterns and thus the optimal values of $\gamma$ significantly differ across models and datasets. 

As shown in Figures~\ref{subfig:biaffine-on-genia}--\ref{subfig:dspert-on-genia}, without negative sampling, the trained model may produce a recall rate much higher than the precision rate. 
Negative sampling dynamically rebalances the precision and recall rates. In general, a higher negative sampling rate $\gamma$ guides the model to classify the within-entity spans more likely as negative samples, which results in a higher precision but a lower recall.\footnote{Empirically, the precision rate also turns to decrease after $\gamma$ exceeds a relatively large value, e.g., 0.1.} Hence, the precision-recall balance can be achieved by setting a good value for $\gamma$. For example, DSpERT finds the optimal $\gamma = 0.005$ on ACE 2004/2005 (Figures~\ref{subfig:dspert-on-ace04} and \ref{subfig:dspert-on-ace05}), and the optimal $\gamma = 0.05$ on GENIA (Figure~\ref{subfig:dspert-on-genia}). On the other hand, in case that the precision and recall are balanced when $\gamma = 0$ (Figures~\ref{subfig:biaffine-on-ace04} and \ref{subfig:biaffine-on-ace05}), negative sampling is unnecessary. 

However, a higher within-entity $F_1$ score from the precision-recall rebalance may not necessarily leads to a higher overall $F_1$ score. Note that (1) $\mathcal{I}$ contains much less ground-truth entities than $\mathcal{O}$, and (2) the predicted entities in $\mathcal{I}$ are always much less precise than those in $\mathcal{O}$. Hence, a relatively high recall in $\mathcal{I}$ has very limited contribution to the overall recall, but yields many false positive samples and thus results in a large drop in the overall precision. On the contrary, high-precision low-recall predicted entities in $\mathcal{I}$ would be a safe and preferred choice. This also explains why \emph{Full Negative} can sometimes achieve high overall $F_1$ scores (e.g., DSpERT on GENIA, Table~\ref{tab:nested-res}). 

Appendices~\ref{app:categorical} and \ref{app:visualization} provide category-specific results and span representation visualizations, respectively.

\section{Results on CoNLL 2003 NFF}
\paragraph{Case Study.} We start from a case study to intuitively demonstrate the results of nested-from-flat NER on the well-known CoNLL 2003 dataset. Specifically, we train DSpERT with \emph{Sampling} ($\gamma = 0.01$) on the training split, and use the trained model to predict entities on the test split. As aforementioned, we have also annotated the nested entities in the test split. 

Table~\ref{tab:case-study} shows 10 example test sentences, marked with the ground-truth and predicted entities. There exist some successful cases that nested entities are correctly recognized. For example, in Sentences~1--4, ``U.S.'', ``Singapore'', ``Melbourne'' and ``Zimbabwe'' are correctly predicted as \texttt{LOC} entities, each within another \texttt{ORG}, \texttt{LOC} or \texttt{MISC} entity; in Sentences~5 and 6, ``Albanian'' and ``Asian'' are correctly recognized as \texttt{MISC} entities, each within another \texttt{ORG} or \texttt{MISC} entity.

\begin{table*}[t]
    \centering \small
    \begin{tabular}{>{\hangindent=2em}p{15.5cm}}
        \toprule
        1. \PER{Mills} is the 38th person to die in \LOC{Florida} 's electric chair since the \ORG{\LOC{U.S.} Supreme Court} reversed itself in 1976 and legalised the death penalty . \\
        \midrule
        2. There is the international prestige \LOC{Singapore} would enjoy , but `` more importantly there is a genuine national interest in fostering better global free trade and an open market '' , said \PER{\ePER{Tan} Kong Yam} , head of Business Policy at the \ORG{National University of \LOC{Singapore}} . \\
        \midrule
        3. \LOC{West Indies} were 53 for two in 15 overs when rain stopped play at the \LOC{\LOC{Melbourne} Cricket Ground} after captain \PER{Courtney \ePER{Walsh}} won the toss and elected to bat . \\
        \midrule
        4. \MISC{\LOC{Zimbabwe} \eMISC{Open}} on Saturday ( \MISC{South \eMISC{African}} unless stated ) \\
        \midrule
        5. \ORG{FIFA} had banned \LOC{Albania} indefinitely after its sports ministry had ordered the suspension of \ORG{\MISC{Albanian} Football Association} general secretary \PER{Eduard Dervishi} and dissolved the executive committee . \\
        \midrule
        6. \LOC{South \eLOC{Korea}} made virtually certain of an \MISC{\MISC{Asian} \eMISC{Cup}} quarter-final spot with a 4-2 win over \LOC{Indonesia} in a Group A match on Saturday . \\
        \midrule
        7. \MISC{Dutch} forward \PER{Reggie \ePER{Blinker}} had his indefinite suspension lifted by \ORG{FIFA} on Friday and was set to make his \ORG{\eORG{\mLOC{Sheffield}} Wednesday} comeback against \ORG{Liverpool} on Saturday . \\
        \midrule
        8. \ORG{\mORG{Bayer} \mLOC{Leverkusen}} ( \LOC{Germany} ) \\
        \midrule
        9. Corrects headline from \ORG{NBA} to \ORG{NHL} and corrects team name in second result from \ORG{\mLOC{La} Clippers} to \ORG{\eORG{\mLOC{Ny}} Islanders} . \\
        \midrule
        10. \LOC{Philadelphia} , which fell from an \eMISC{\eMISC{NFC} \eMISC{East}} tie with the \ORG{\eORG{\mLOC{Dallas}} \eORG{Cowboys}} and \ORG{\eORG{\mLOC{Washington}} \eORG{Redskins}} , go on the road against the \ORG{\mLOC{New York} \eORG{Jets}} and then entertain \eLOC{\mORG{Arizona}} . \\
        \bottomrule
    \end{tabular}
    \caption{Example sentences with ground-truth and predicted entities from the test split of CoNLL 2003 NFF. 
    The \textcolor{casgreen}{green} entities are true positive samples, the \textcolor{casred}{red} ones are false negative, and the \textcolor{casyellow}{orange} ones are false positive.}
    \label{tab:case-study}
\end{table*}

The incorrect recognition results contain the following typical scenarios:
\begin{itemize}
    \item The first or last name within a full person name, as a false positive \texttt{PER} entity (e.g., Sentences~2, 3, 7). 
    \item A geopolitical concept within a specific location name, as a false positive \texttt{LOC} or \texttt{MISC} entity (e.g., Sentence~4, 6). 
    \item The anchor word of an event/organization name, as a false positive \texttt{MISC}/\texttt{ORG} entity (e.g., Sentences~4, 6, 10). 
    \item A \texttt{LOC} entity mislabeled as \texttt{ORG} within an organization name (e.g., Sentences~7, 9, 10). 
    \item Nested entities that rarely appear independently at the topmost level in the corpus (e.g., Sentences~8, 10), especially for abbreviations (e.g., Sentence~9). 
\end{itemize}

Most scenarios are largely attributable to the \emph{annotation inconsistency}, i.e., the inconsistency of the annotation standards between the nested and flat entities. For example, (1) if an entity mention is nested within its full name in text, nested NER annotation guidelines~\citep[e.g., ACE;][]{doddington-etal-2004-automatic} typically label the full name only, but ignore the substring mention. This avoids redundancy, since the two mentions refer to a same entity concept. However, the same substring should be annotated if it appears at the topmost level in text. (2) CoNLL 2003 contains a large amount of sports news, where city/country names are ubiquitously used to refer to team names; such mentions should be annotated as \texttt{ORG} entities according to the guidelines~\citep{tjong-kim-sang-de-meulder-2003-introduction}. However, the same city/country mentions should be annotated as \texttt{LOC} entities if they appear within the full team names. Given such inconsistencies, a model trained by flat supervision plausibly learns patterns inapplicable to the nested entities. This results in redundant or mislabeled entities, although some of them might be acceptable in practice. 

Some scenarios are associated with the \emph{data inconsistency}, i.e., the inconsistency of the data distributions between the within-entity and out-of-entity spans. Some nested entity mentions almost never appear independently at the topmost level in text. For example, some location abbreviations (e.g., ``NY'' or ``LA'') are always nested within other entities in the corpus. This poses a very challenging case for the nested-from-flat NER task, due to the lack of supervision. Actually, a nested-from-flat NER model may never succeed in that case unless sufficient external knowledge is introduced and utilized, such as knowledge databases~\citep{wang-etal-2021-improving, geng2022tuner} or more powerful PLMs.

\paragraph{Post Processing.} Based on the above analysis, we propose two post-processing operations on the predicted entity set: 
\begin{itemize}
    \item If a \texttt{PER} entity is nested within another \texttt{PER} entity, remove the nested one; because it is probably a first/last name inside the full name. 
    \item If an \texttt{ORG} entity is nested within another entity, change the entity label to \texttt{LOC}; because it is probably a location name used to refer to team names in other context in the corpus. 
\end{itemize}

\begin{table}[t]
    \centering \small
    \begin{tabular}{lccc}
        \toprule
         & Within & Out & Overall \\
        \midrule
        Full Negative            & \avestd{14.6}{0.2} & \avestd{\textbf{93.4}}{0.1} & \avestd{\textbf{89.8}}{0.1} \\
        \quad w/ Post Processing & \avestd{14.6}{0.2} & -- & \avestd{\textbf{89.7}}{0.1} \\
        Sampling                 & \avestd{16.6}{1.1} & \avestd{\textbf{93.5}}{0.2} & \avestd{84.3}{0.6} \\
        \quad w/ Post Processing & \avestd{\textbf{33.2}}{1.0} & -- & \avestd{88.6}{0.3} \\
        Full Ignoring            & \avestd{~~8.7}{0.4} & \avestd{92.9}{0.4} & \avestd{70.3}{0.9} \\
        \quad w/ Post Processing & \avestd{31.8}{1.2} & -- & \avestd{82.4}{0.8} \\
        \bottomrule
    \end{tabular}
    \caption{Results of nested-from-flat experiments by DSpERT on CoNLL 2003 NFF. Reported are average $F_1$ scores with corresponding standard deviations of 10 independent runs. The best $F_1$ scores are in bold.}
    \label{tab:conll2003-res}
\end{table}

Table~\ref{tab:conll2003-res} lists the evaluation results of DSpERT on CoNLL 2003 NFF. With the help of post processing, DSpERT+\emph{Sampling} achieves the best within-entity $F_1$ score of 33.3\%. This score seems low, relative to those on ACE 2004/2005 (i.e., 54\%+). The main reason is that CoNLL 2003 specifies that named entities should be unique identifiers like proper names or acronyms~\citep{tjong-kim-sang-de-meulder-2003-introduction}, while ACE additionally includes pronouns or descriptions that refer to entities~\citep{doddington-etal-2004-automatic}. Note that the pronouns and descriptions can be labeled more consistently between the within-entity and out-of-entity spans, which lowers the difficulty of nested-from-flat NER on ACE 2004/2005. In other words, CoNLL 2003 NFF poses a more strict and challenging benchmark of nested-from-flat NER. 

Similar to the results on other datasets, our model is trained to recognize nested entities without affecting the out-of-entity performance. Hence, compared to a model that predicts flat entities only, our method always has merit for the additional ability of nested entity recognition.

\section{Discussion and Conclusion}
Although the NLP community has undertaken increasing efforts to investigate and develop nested NER models, many existing NER resources are flatly designed and annotated, especially in languages other than English. For example, the widely-used Chinese NER benchmarks, e.g., OntoNotes 4, MSRA~\citep{levow-2006-third}, Weibo NER~\citep{peng-dredze-2015-named} and Resume NER~\citep{zhang-yang-2018-chinese}, are all flat; similar situation holds for Japanese~\citep{iwakura-etal-2016-constructing}, Korean~\citep{jeong-etal-2020-constructing}, Vietnamese~\citep{truong-etal-2021-covid}, etc. Most domain-specific entity recognition datasets are also designed in a flat scheme~\citep{uzuner2011i2b2, albright2013towards, jeong-etal-2020-constructing}. 

Nested-from-flat NER corresponds to a realistic application scenario: given training data annotated with flat entities only, one may still desire the trained model capable of recognizing nested entities. This task is theoretically feasible, because the recognizable patterns for outermost entities should be, at least partially, transferable to nested entities. To the best of our knowledge, this study is the first to validate and investigate this mechanism. 

On the other hand, nested-from-flat NER is a challenging setting because of the data and annotation inconsistencies between the within-entity and out-of-entity spans. Hence, the models may learn inapplicable or insufficient patterns when transferred to recognizing nested entities. 

We choose the span-based NER framework because it explicitly distinguishes between positive and negative spans, which allows us to flexibly manipulate the negative samples in the within-entity area. Since the within-entity spans probably contain unlabeled nested entities, it is straightforward to ignore these spans in loss computation; while we empirically find it beneficial to apply negative sampling with a very small rate (i.e., 0.01). The negative sampling is inspired by \citet{li2021empirical}'s solution for unlabeled entity problem, but their optimal sampling rate is much larger (0.3 -- 0.4).

In conclusion, this study proposes nested-from-flat NER, a new subtask that asks to train a nested NER model with flatly annotated data. We find a simple but effective solution to this task. With nested entities removed from the training data, our model can achieve 54.8\%, 54.2\% and 41.1\% within-entity $F_1$ scores on ACE 2004, ACE 2005 and GENIA, respectively. Moreover, the model's performance on flat entity recognition is completely unaffected by its additional ability to recognize nested entities. We further propose a nested-from-flat NER benchmark, CoNLL 2003 NFF, which consists of CoNLL 2003 and our annotations of nested entities in the test set. With in-depth case study, we find that the main challenges stem from the data and annotation inconsistencies between the flat and nested entities. 

Future work may improve the nested-from-flat NER performance by utilizing more external knowledge, either explicitly via knowledge databases~\citep{wang-etal-2021-improving, geng2022tuner} or implicitly with more powerful PLMs.


\bibliography{anthology,custom,references}
\bibliographystyle{acl_natbib}

\newpage
\appendix

\section{Datasets} \label{app:datasets}
\paragraph{ACE 2004 and ACE 2005} are two English nested NER datasets created by the Automatic Content Extraction (ACE) Program~\citep{doddington-etal-2004-automatic}. The corpus consists of broadcast transcripts, newswire and newspaper data; the entity types include Person (\texttt{PER}), Organization (\texttt{ORG}), Facility (\texttt{FAC}), Location (\texttt{LOC}), Geo-political Entity (\texttt{GPE}), Vehicle (\texttt{VEH}), and Weapon (\texttt{WEA}). Our data processing and splits follow \citet{lu-roth-2015-joint}. 

As indicated by the annotation guidelines, ACE aims to recognize all mentions of entities, not just names. In other words, an entity mention can be a name, a description, or a pronoun, as long as it clearly refers to the entity.

\paragraph{GENIA} is a nested NER dataset on English biological articles~\citep{kim2003genia}. There are five entity categories, i.e., DNA, RNA, Protein, Cell Line, and Cell Type. Our data processing follows \citet{lu-roth-2015-joint}, and data splits follow \citet{yan-etal-2021-unified-generative} and \citet{li2022unified}. 

\paragraph{CoNLL 2003 NFF} is a nested-from-flat NER benchmark that consists of the text data and flat NER annotations of CoNLL 2003~\citep{tjong-kim-sang-de-meulder-2003-introduction}, and our annotations of nested entities in the test split. The corpus consists of Reuters news stories in 1996 and 1997; the entity categories are Person (\texttt{PER}), Organization (\texttt{ORG}), Location (\texttt{LOC}), and Miscellaneous (\texttt{MISC}). We use the original data splits for experiments. 

In CoNLL 2003, named entities are limited to unique identifiers, such as proper names and acronyms. It excludes pronouns and descriptions that refer to entities. 

We hired three NLP experts to additionally annotate the nested entities in the test split. BRAT Rapid Annotation Tool~\citep{stenetorp2012brat} was deployed to provide the annotation user interface. The annotators were asked to carefully read through the CoNLL 2003 Annotation Guidelines, and strictly follow the guidelines when manually labeling the nested entities that had been ignored in the original data. All the original annotations, even a few incorrect ones~\citep{wang-etal-2019-crossweigh}, are retained without modification. Each annotator labeled all documents in the test set, and the final results are based on an additional round of manual validation that resolves the inter-annotator disagreements. 

Table~\ref{tab:descriptive} presents the descriptive statistics of the datasets. 

\begin{table*}[t]
    \centering \small
    \begin{tabular}{lrrrrrrrrrrrr}
        \toprule
         & \multicolumn{3}{c}{ACE 2004} & \multicolumn{3}{c}{ACE 2005} & \multicolumn{3}{c}{GENIA} & \multicolumn{3}{c}{CoNLL 2003 NFF} \\
        \cmidrule(lr){2-4} \cmidrule(lr){5-7} \cmidrule(lr){8-10} \cmidrule(lr){11-13} 
         & Train & Dev. & Test & Train & Dev. & Test & Train & Dev. & Test & Train & Dev. & Test \\
        \midrule
        \#Sentence        & 6,799  & 829   & 879   & 7,336  & 958   & 1,047 & 15,023 & 1,669 & 1,854 & 14,987 & 3,466 & 3,684 \\
        \quad Nested (\%) & 39.5   & 35.3  & 42.4  & 36.6   & 35.6  & 31.5  & 21.3   & 19.5  & 24.1  & --     & --    & 11.3 \\
        \#Entity          & 22,207 & 2,511 & 3,031 & 24,687 & 3,217 & 3,027 & 46,164 & 4,371 & 5,511 & 23,499 & 5,942 & 5,648 \\
        \quad Nested (\%) & 28.2   & 27.2  & 29.1  & 24.4   & 22.2  & 23.8  & 9.5    & 9.6   & 11.4  & --     & --    & 7.9 \\
        \quad Ave. Len.   & 2.5    & 2.6   & 2.5   & 2.3    & 2.1   & 2.3   & 1.9    & 2.1   & 2.1   & 1.4    & 1.4   & 1.4 \\
        \quad Max. Len.   & 57     & 35    & 43    & 49     & 30    & 27    & 17     & 18    & 15    & 10     & 10    & 6 \\
        \bottomrule
    \end{tabular}
    \caption{Descriptive statistics of datasets. ``\#Sentence'' denotes the number of sentences, under which ``Nested (\%)'' denotes the proportion of sentences with nested entities. ``\#Entity'' denotes the number of entities, under which ``Nested (\%)'' denotes the proportion of nested entities, ``Ave. Len.'' and ``Max. Len.'' denote the average and maximum lengths of entities, respectively.}
    \label{tab:descriptive}
\end{table*}

\section{Category-Specific Results} \label{app:categorical}
Table~\ref{tab:categorical} lists the categorical results of within-entity $F_1$ scores on ACE 2004, ACE 2005 and GENIA. 

The performance significantly varies across categories. Specifically, the categorical $F_1$ scores range from 38\% to 60\% on ACE 2004/2005, from 18\% to 51\% on GENIA. In general, the model performs better on categories that contain more entities, such as \texttt{PER} and \texttt{GPE} in ACE 2004/2005, and the Protein type in GENIA. This is reasonable, since such categories have more positive samples in the training data, which enable the model to learn more accurate decision boundaries. 

The categorical $F_1$ scores by nested-from-flat models are consistently lower than, and positively correlated with the corresponding scores by gold supervision. The Pearson correlation coefficients between the categorical $F_1$ scores are positive for all the three datasets. One exception is the Cell Type category in GENIA, where the nested-from-flat model surprisingly outperforms its counterpart with gold supervision; we conjecture that some Cell Type entities are incorrectly annotated in the training data and thus misguide the trained model. 

\begin{table}[t]
    \centering \small
    \begin{tabular}{ccccc}
        \toprule
         & \multicolumn{2}{c}{ACE 2004} & \multicolumn{2}{c}{ACE 2005} \\
        \cmidrule(lr){2-3} \cmidrule(lr){4-5}
         & Sampling & \gstyle{Gold S.} & Sampling & \gstyle{Gold S.} \\
        \midrule
        \texttt{PER} & \avestd{55.7}{0.8} & \gavestd{90.0}{0.3} & \avestd{57.9}{0.8} & \gavestd{89.6}{0.4} \\
        \texttt{ORG} & \avestd{48.9}{1.6} & \gavestd{81.8}{0.5} & \avestd{49.5}{1.6} & \gavestd{80.7}{1.2} \\
        \texttt{FAC} & \avestd{51.0}{3.3} & \gavestd{82.4}{1.9} & \avestd{52.9}{4.5} & \gavestd{78.8}{2.1} \\
        \texttt{LOC} & \avestd{43.7}{3.6} & \gavestd{77.4}{1.5} & \avestd{37.8}{3.2} & \gavestd{75.0}{3.5} \\
        \texttt{GPE} & \avestd{59.4}{2.6} & \gavestd{88.6}{0.5} & \avestd{57.9}{4.3} & \gavestd{88.4}{1.0} \\
        \texttt{VEH} & \avestd{37.9}{7.3} & \gavestd{85.7}{0.0} & \avestd{46.7}{2.0} & \gavestd{81.4}{3.2} \\
        \texttt{WEA} & \avestd{55.8}{8.2} & \gavestd{75.2}{2.4} & \avestd{38.4}{1.7} & \gavestd{69.6}{3.0} \\
        \midrule
        Correlation  & \multicolumn{2}{c}{0.203} & \multicolumn{2}{c}{0.909} \\
        \bottomrule 
        \toprule
         & \multicolumn{2}{c}{GENIA} \\
        \cmidrule(lr){2-3} 
         & Sampling & \gstyle{Gold S.} \\
        \midrule
        DNA          & \avestd{22.9}{1.4} & \gavestd{33.0}{1.4} \\
        RNA          & --                 & -- \\
        Protein      & \avestd{51.0}{0.7} & \gavestd{65.3}{0.8} \\
        Cell Line    & \avestd{17.9}{2.5} & \gavestd{28.3}{2.3} \\
        Cell Type    & \avestd{28.1}{1.2} & \gavestd{11.6}{1.3} \\
        \midrule
        Correlation  & \multicolumn{2}{c}{0.788} \\
        \bottomrule
    \end{tabular}
    \caption{Categorical within-entity $F_1$ scores by DSpERT on nested NER datasets. Reported are average $F_1$ scores with corresponding standard deviations of 10 independent runs. 
    ``\gstyle{Gold S.}'' indicates results with gold supervision.
    ``--'' means no ground-truth nested entities in the test set.}
    \label{tab:categorical}
\end{table}

\section{Visualization of Span Representations} \label{app:visualization}
\begin{figure*}[t]
    \centering
    \begin{subfigure}{0.45\textwidth}
    \centering
    \includegraphics[width=\textwidth]{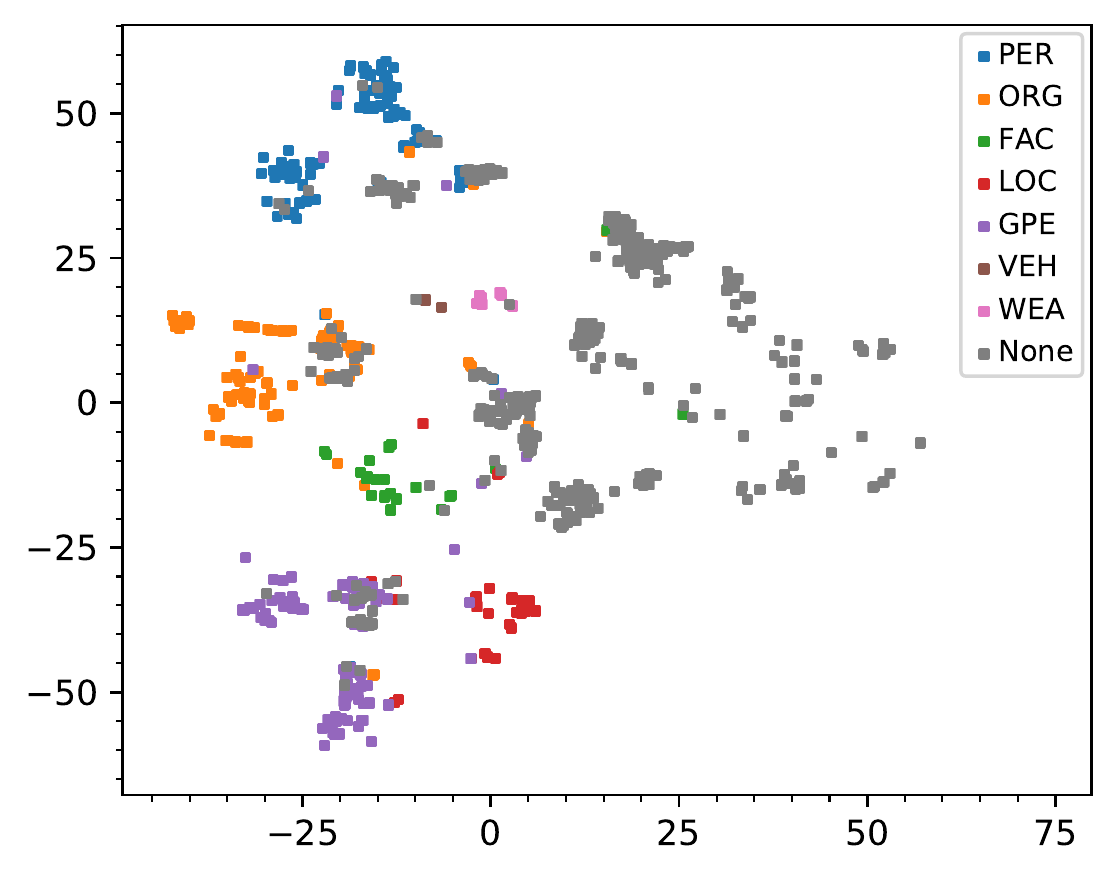}
    \caption{Sampling, Within-entity} \label{subfig:sampling-on-within}
    \end{subfigure}
    \begin{subfigure}{0.45\textwidth}
    \centering
    \includegraphics[width=\textwidth]{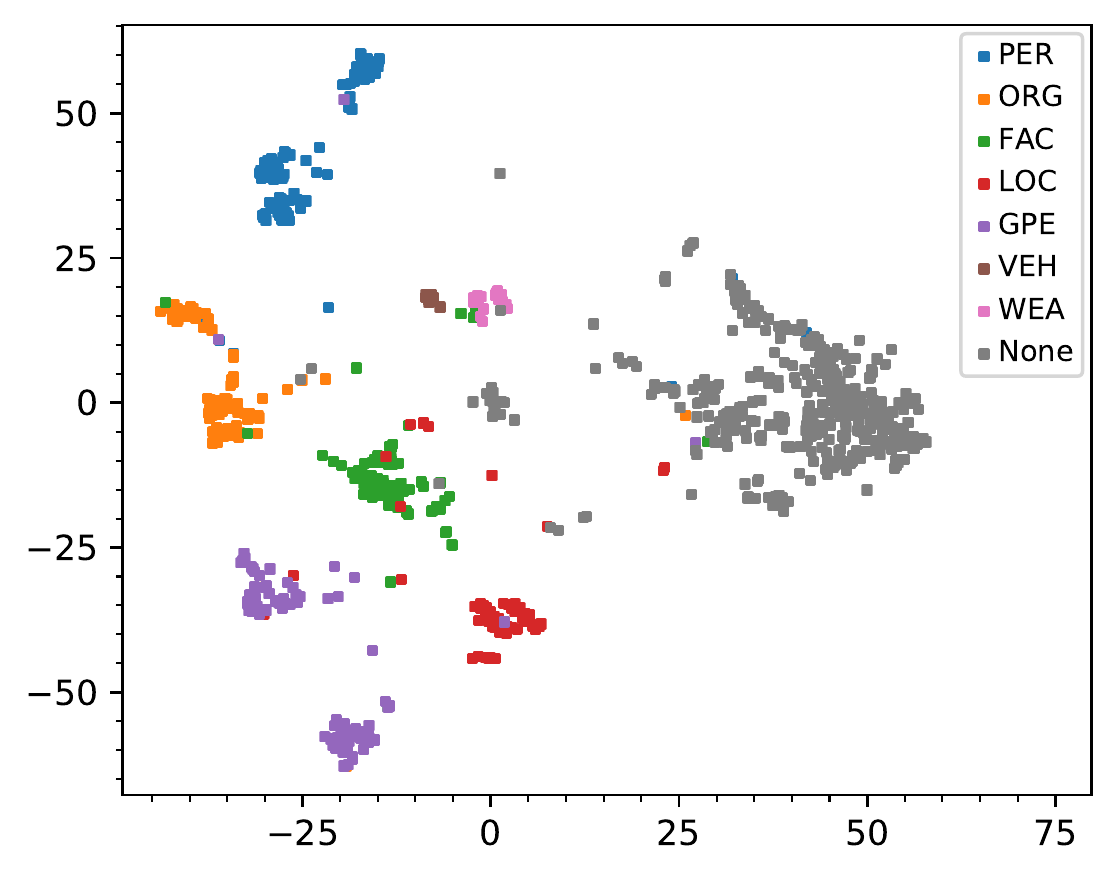}
    \caption{Sampling, Out-of-entity} \label{subfig:sampling-on-out}
    \end{subfigure}
    \begin{subfigure}{0.45\textwidth}
    \centering
    \includegraphics[width=\textwidth]{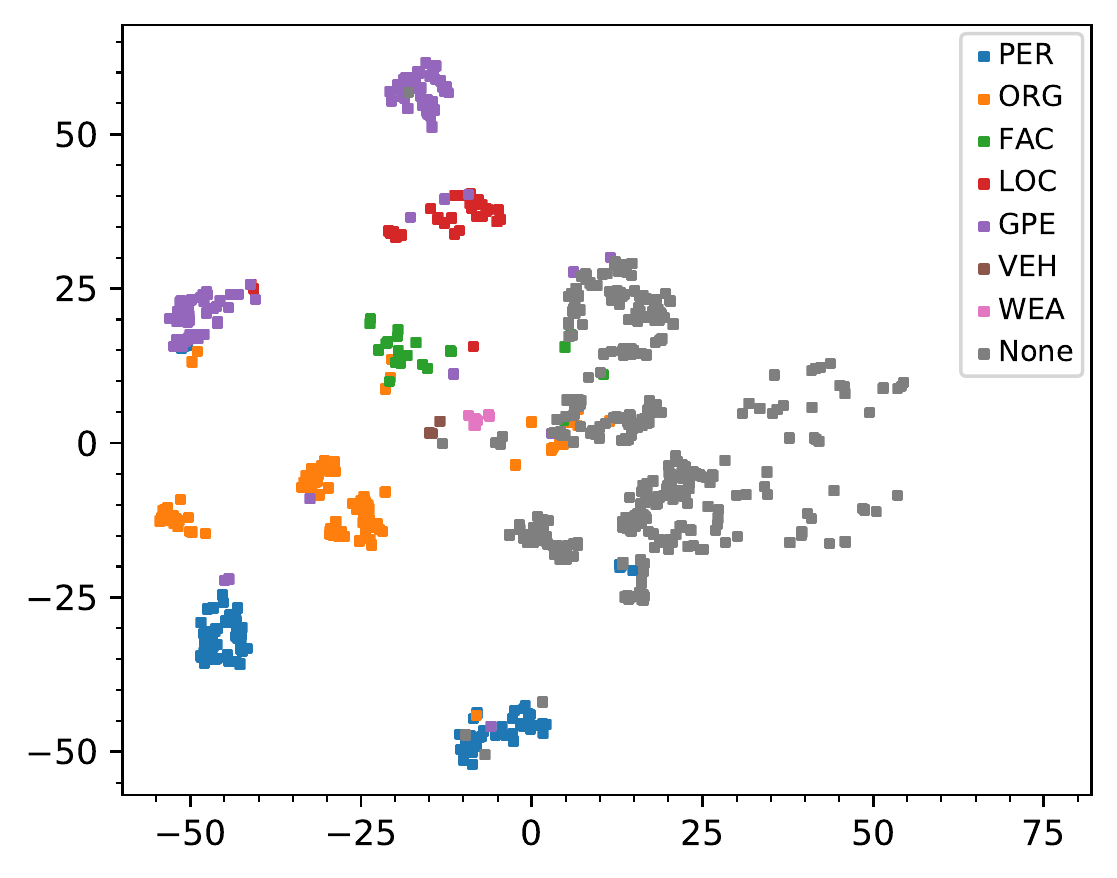}
    \caption{\gstyle{Gold Superv.}, Within-entity} \label{subfig:gold-on-within}
    \end{subfigure}
    \centering
    \begin{subfigure}{0.45\textwidth}
    \centering
    \includegraphics[width=\textwidth]{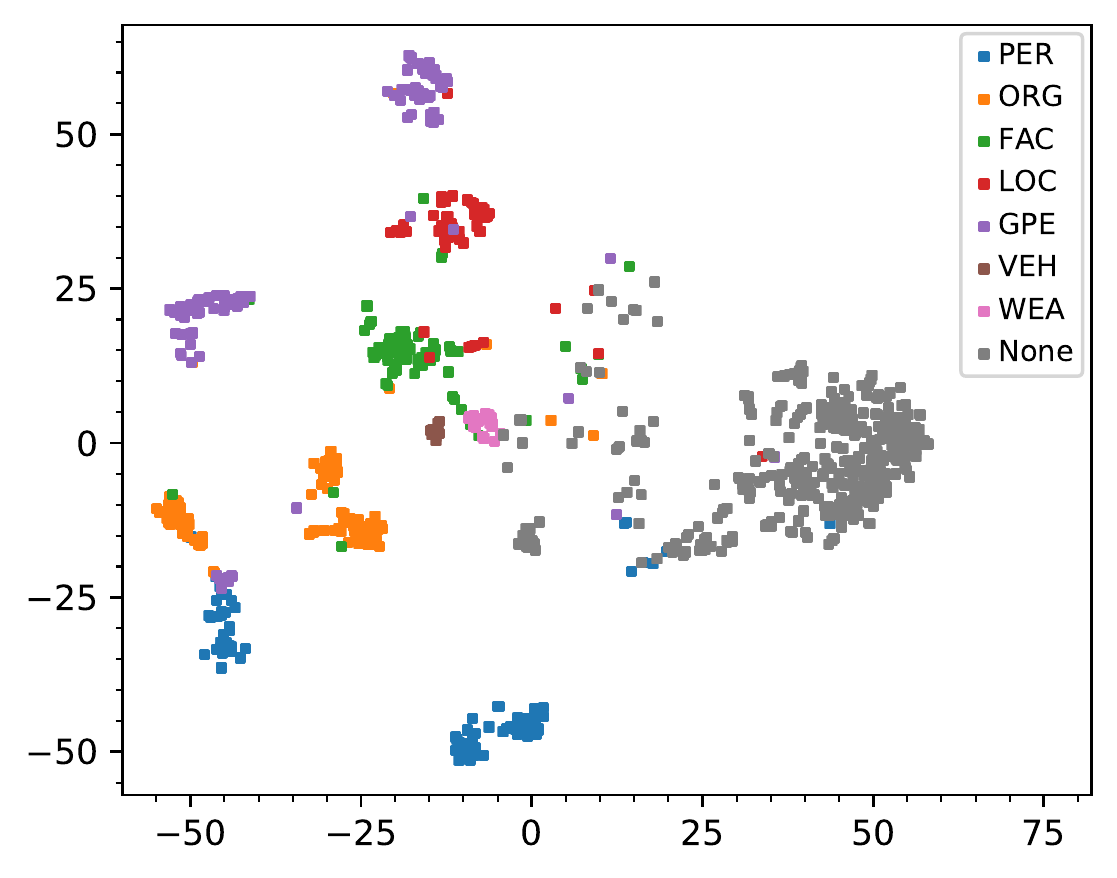}
    \caption{\gstyle{Gold Superv.}, Out-of-entity} \label{subfig:gold-on-out}
    \end{subfigure}
    \caption{t-SNE visualization of pre-logit span representations by DSpERT on ACE 2004 test sentences. 
    Each row compares the within-entity and out-of-entity span representations from a same model, visualized by a shared t-SNE.}
    \label{fig:tsne}
\end{figure*}

Figure~\ref{fig:tsne} presents the t-SNE visualizations~\citep{van2008visualizing} of the pre-logit span representations. The representations are constructed by DSpERT on the test sentences of ACE 2004. 

For the model trained by flat supervision, the within-entity span representations are largely clustered by categories, but a part of negative samples are mixed into the positive clusters, resulting in unclear and ambiguous decision boundaries (Figure~\ref{subfig:sampling-on-within}). In contrast, the out-of-entity span representations form clear and tight categorical clusters (Figure~\ref{subfig:sampling-on-out}). However, if the model is trained on data with nested annotations, both the within-entity and out-of-entity representations are clearly clustered by categories (Figures~\ref{subfig:gold-on-within}, \ref{subfig:gold-on-out}). 

Hence, the spans mixed across positive and negative clusters in Figure~\ref{subfig:sampling-on-within} lack supervision. As suggested by the case study on CoNLL 2003 NFF, these span samples probably correspond to the data and annotation inconsistencies between the within-entity and out-of-entity spans. They are particularly difficult to discriminate in the nested-from-flat setting.

\end{document}

%% file: tikz/inside-and-outside-spans.tex
\begin{tikzpicture}
    \foreach \i in {0,1,...,10}{
        \draw[dotted] (0, -\i) -- (10, -\i);
        \draw[dotted] (\i, 0) -- (\i, -10);
    }
    \node[left, font=\sffamily\Large, rotate=90, anchor=center] at (-0.8, -5) {Start};
    \node[below, font=\sffamily\Large] at (5, -11.6) {End};
    \foreach \i/\w in {0/Mr., 1/John, 2/Smith, 3/graduated, 4/from, 5/New, 6/York, 7/University, 8/last, 9/year}{
         \node[left, font=\sffamily\Large] at (0, -\i-0.5) {\i};
         \node[below, anchor=east, rotate=30, font=\sffamily] at (\i+1, -10.1) {\w};
         \node[below, font=\sffamily\Large] at (\i+0.5, -11) {\i};
    }
    \foreach \i in {0,1,...,9}{
        \foreach \j in {\i,...,9}{
            \draw[rounded corners, fill=gray!30] (\j, -\i) rectangle (\j+1, -\i-1);
        }
    }
    \foreach \i in {1,2}{
        \foreach \j in {\i,2}{
            \draw[rounded corners, fill=casblue!50] (\j, -\i) rectangle (\j+1, -\i-1);
        }
    }
    \foreach \i in {5,...,7}{
        \foreach \j in {\i,...,7}{
            \draw[rounded corners, fill=casblue!50] (\j, -\i) rectangle (\j+1, -\i-1);
        }
    }
    \draw[rounded corners, fill=casred!80] (2, -1) rectangle (3, -2);
    \draw[rounded corners, fill=casred!80] (7, -5) rectangle (8, -6); 
    \draw[rounded corners, pattern=north east lines, pattern color=casred!80] (6, -5) rectangle (7, -6);
\end{tikzpicture}